\documentclass{article}

\usepackage[preprint]{neurips_2023}

\usepackage[utf8]{inputenc} 
\usepackage[T1]{fontenc}    
\usepackage{xcolor}         
\definecolor{myblue}{rgb}{0.0 0.0 0.5}
\usepackage{hyperref}       
\hypersetup{
  colorlinks=true,
  linkcolor=myblue,
  citecolor=myblue,
  filecolor=myblue,
  urlcolor=myblue,
}
\usepackage{url}            
\usepackage{booktabs}       
\usepackage{amsfonts}       
\usepackage{nicefrac}       
\usepackage{microtype}      
\usepackage{paralist}

\newcommand{\addcite}[1]{\textcolor{red}{[add cite]}}

\title{Responsible AI Research Needs Impact Statements Too}

\author{%
  Alexandra~Olteanu \\
  Microsoft Research \\
  \And
  Michael Ekstrand \\
  Drexel University
  \And
  Carlos Castillo \\
  ICREA and UPF
  \And
  Jina Suh \\
  Microsoft Research
}

\begin{document}

\maketitle

All types of research, development, and policy work can have unintended, adverse consequences---work in responsible artificial intelligence (RAI), ethical AI, or ethics in AI is {\em no} exception.

The work of the responsible AI community has illustrated how the design, deployment, and use of computational systems---including machine learning (ML), artificial intelligence (AI), and natural language processing (NLP) systems---engender a range of adverse impacts. 
As a result, in recent years, the authors of ML, AI, and NLP research papers have been required to include reflections on possible unintended consequences and negative social impacts as dedicated sections or extensive checklists \citep[see e.g., ][]{ashurst2022ai,nanayakkara2021unpacking,boyarskaya2020overcoming,prunkl2021institutionalizing}.
Even though such requirement traces its roots to work done within the responsible AI community \citep[e.g., ][]{hecht2021s}, responsible AI conferences and publication venues, such as FAccT,\footnote{ACM Conference on Fairness, Accountability, and Transparency \url{https://facctconference.org/}} AIES,\footnote{AAAI/ACM Conference on Artificial Intelligence, Ethics, and Society \url{https://www.aies-conference.com}} FORC,\footnote{Symposium on Foundations of Responsible Computing \url{https://responsiblecomputing.org}} or EAAMO,\footnote{ACM Conference on Equity and Access in Algorithms, Mechanisms, and Optimization \url{https://eaamo.org}} have yet to explicitly enforce similar requirements.
~Surprisingly, many papers on responsible AI, ethical AI, ethics in AI, or related topics\footnote{Throughout this viewpoint, we use the acronym RAI to broadly refer to work on responsible AI/computing, ethical AI/computing, trustworthy AI/computing, ethics in AI/computing, or any related topics.} do not include similar reflections on possible adverse impacts. RAI research and work is often taken to be inherently beneficial with little to no potential for harm and can thus paradoxically fail to consider any possible adverse consequences it may give rise to~\citep{boyarskaya2020overcoming}. This is also the case for many RAI artifacts which were found to, e.g., ``not contend with the organizational, labor, and political implications of AI ethics work in practice''~\citep{wong2023seeing}.

This trend of failing to reflect on the possible negative impact of our own work should concern all of us, as the research we conduct and the artifacts we build are more often than not {\em value-laden}, and thus encode all kinds of implicit practices, assumptions, norms, and values \citep[e.g., ][]{jakesch2022different,raji2022fallacy,craftTOC2023,pinneyMuchAdoGender2023,zhou2022deconstructing}. Similarly to our colleagues from other research communities, we---RAI researchers and practitioners---can and often do also suffer from similar ``failures of imagination'' when it comes to the impact of our own work, and we need to {\em at least} keep ourselves to the same standard that we expect other communities to adhere to. 

We believe {\em responsible AI research needs impact statements, too}.

\paragraph{Requiring adverse impact statements for RAI research is long overdue.} 
There have been growing concerns about how our own work has routinely failed to engage with and address deeper patterns of injustice and inequality, often assuming that many elements of the status quo are immutable~\citep{keyes2019mulching,green2020algorithmic,abebe2020roles,laufer2022four}.  
We know that common RAI values may conflict in certain deployment settings and that different groups assess and prioritize responsible AI values differently~\citep{jakesch2022different}, with RAI research still largely centering mostly Westernized and  US-centric perspectives~\citep{septiandri2023weird}. All these can have profound implications for what problems and solutions end up being prioritized. 

Even well-intentioned applications, policies, or interventions to mitigate known issues can and often do lead to harm, e.g.,~\citep{green2020algorithmic}.
Scrutiny is required even when a system, practice or framework seems to address real needs stakeholders might have, as there can be subtle patterns of problematic uses, system behavior, or outcomes that might be harder to discern~\citep{sandvig2014auditing,robertson2021can,olteanu2019social}. 
For instance, \cite{bennett2020point} discuss how fixating on certain notions of fairness can reinforce existing dynamics and exacerbate harms. 
Indeed, blindly adhering to some RAI frameworks without considering what exactly are we trying to make, e.g., fair or transparent, can lead to these frameworks being used to legitimize harmful, absurd technologies~\citep{keyes2019mulching} and to a ``checkbox culture''~\citep{balayn2023fairness} where researchers and practitioners do not meaningfully engage with RAI considerations or the social, economic, and political origins of these considerations.

Furthermore, a focus on bias and fairness claims often assumes that these issues are due to poor implementation of a system and center the algorithmic systems themselves, distracting from both basic validation of functionality~\citep{raji2022fallacy} and the factors that led to injustices in the first place~\citep{bennett2020point}. RAI research, similarly to much of AI research, inadvertently may take for granted that AI systems work or that they are inevitable~\citep{raji2022fallacy}, failing to reflect on whether techno-solutions are even justifiable. 
Similarly, RAI interventions targeting the design phase of AI life-cycle tend to ignore important contextual factors that determine the outcomes resulting from the implementation, deployment, and use of AI systems~\citep{gansky2022counterfacctual}. 
This is because many algorithms developed to help guarantee various, e.g., ``fairness'' requirements are developed ``without policy and societal contexts in mind.''\footnote{\url{https://www.wired.com/story/bias-statistics-artificial-intelligence-healthcare/}}

There are also concerns about how RAI practice and research risks to facilitate paying lip service to the issues it ostensibly aims to address \citep[e.g., ][]{ali2023walking}, rather than driving meaningful changes. 
RAI work might ignore organizational power dynamics and structures~\citep{wong2023seeing,ali2023walking} that are critical to enacting change, as well as the fact that in practice the responsibility of doing this work many times falls on the shoulders of either individuals coming from marginalized backgrounds~\citep{birhane2022forgotten} and/or on those of time-constrained and untrained practitioners~\citep{rakova2021responsible,buccinca2023aha}. 
Raising concerns and performing RAI work can also take a psychological toll on RAI practitioners~\citep{widder2023s,heikkila2022responsible} as, e.g., they might be exposed to harmful content or might need to take great personal risks. 

{\em Examples of how RAI research and work can thus also inadvertently lead to harmful outcomes abound.}  

\paragraph{What are other research communities doing?} 
Following the call by \citet{hecht2021s} for researchers to disclose possible negative consequences of their work, conferences like the Conference on Neural Processing Information Systems (NeurIPS)~\citep{ashurst2020guide} and the International Conference on Machine Learning (ICML) have started requiring authors to reflect on ``{whenever there are risks associated with the proposed methods, methodology, application or data collection and data usage, authors are expected to elaborate on the rationale of their decision and potential mitigations.}''\footnote{NeurIPS Code of Ethics: \url{https://neurips.cc/public/EthicsGuidelines}}
These requirements have evolved over time from dedicated statements on  ``potential broader impact of their work, including its ethical aspects and future societal consequences''~\citep{ashurst2020guide} to a detailed paper checklist.\footnote{NeurIPS Paper Checklist Guidelines: \url{https://neurips.cc/public/guides/PaperChecklist}}
%
Similarly, the International Conference on Learning Representations (ICLR) also encourages authors to include an Ethics Statement in their papers that covers reflections about ``potentially harmful insights, methodologies and applications.''\footnote{Author guide for the International Conference on Learning Representations: \url{https://iclr.cc/Conferences/2024/AuthorGuide}}
The current Association for Computational Linguistics (ACL) rolling review call for papers---used by most ACL venues---explicitly encourages the authors ``to discuss the limitations of their work in a dedicated section'' and ``devote a section of their paper to concerns about the ethical impact of the work and to a discussion of broader impacts of the work,''\footnote{\url{https://aclrollingreview.org/cfp}} while also providing a responsible NLP research checklist.\footnote{\url{https://aclrollingreview.org/responsibleNLPresearch/}}
The conference on Empirical Methods in Natural Language Processing (EMNLP) made the ``discussion of limitations'' mandatory in 2023, while also encouraging authors to include ``an optional broader impact statement or other discussion of ethics.''\footnote{\url{https://2023.emnlp.org/calls/main_conference_papers/}}
To nudge authors to be comprehensive in their discussions of limitations, ethical considerations, and adverse impacts, these venues typically do not count these sections or discussions towards the page limit.


\paragraph{What do RAI venues do?} 
The ACM Conference on Fairness, Accountability, and Transparency (FAccT) recent calls for papers have mainly guided authors towards the new ACM Code of Ethics and Professional Conduct,\footnote{The ACM Code of Ethics and Professional Conduct notes that computing professionals' responsibilities include: ``Give comprehensive and thorough evaluations of computer systems and their impacts, including analysis of possible risks.'' and ``Foster public awareness and understanding of computing, related technologies, and their consequences.''} and ask them to ``adhere to precepts of ethical research and community norms.''
Similarly, past call for papers of other RAI venues such as FORC and EEAMO only briefly note that authors e.g., ``are encouraged to reflect on relevant ethics guidelines'' such as the ACM Code of Ethics, respectively that ``papers should include a discussion of ethical impacts and precautions taken, including disclosure regarding whether the study was approved by an Institutional Review Board (IRB).'' 
AIES' call for papers\footnote{\url{https://www.aies-conference.com/2022/call-for-papers/index.html}} does not seem to include any language requiring or encouraging papers to include ethical considerations, limitations, or impact statements.
%
Overall, these CFPs do not feature explicit calls for authors to include reflections on possible adverse impacts their work might give rise to, do not explicitly enforce such requirements, and do not provide explicit guidance or incentives to do so (e.g., extra pages, checklists).

\section*{Suggestions for More Meaningful Engagement with the Impact of RAI Research}

To help others understand not only the benefits or positive outcomes, but also the possible harmful outcomes or adverse impacts of our own research, we believe RAI papers should go one step beyond  what other research communities are currently doing and include: \begin{inparaenum}[\bf 1)]
\item reflections on how the researchers' disciplinary background, lived experiences, and goals might affect the way they approach their work (as part of {\em researcher positionality statements}), 
\item a description of the ethical concerns the authors grappled with and mitigated while conducting the work (as part of {\em ethical considerations statements}), 
\item reflections on the limitations of their methodological choices (as part of a {\em discussion of limitations}), 
and---informed by a researcher positionality, known ethical concerns, and known limitations---\item reflections on possible adverse impacts the work might lead to once published (as part of {\em adverse impact statements}).
\end{inparaenum}

By distinguishing between these four different elements of research practice and outcomes---which have at times been conflated---without being too prescriptive, we hope to provide both some clarity and guidance about what each of these statements could include. 
In doing so, we draw on emerging practices in other communities \citep[e.g.,][]{ashurst2020guide,hecht2020suggestions}.
However, we recognize that the RAI community comes from diverse disciplinary backgrounds, and some of these elements might be unfamiliar or might be less applicable for some types of work than others.



\paragraph{1) RAI papers should include {\em researcher positionality statements}.} 
Our research, development, and policy work necessarily rely on various (explicit and implicit) assumptions that we make and that are being shaped by our values, disciplinary backgrounds, knowledge, and lived experiences. 
We collectively hold a variety of goals and theories of change that motivate and guide our work \citep{craftTOC2023}. 
Positionality statements are meant to provide added transparency and scaffold readers' understanding of how our background and experiences influence or inform our work, and how our perspectives might as a result differ from those of others~\citep{liang2021embracing}. 
If the authors believe that their worldview does not affect their work, that by itself reflects a position that the authors operate under, and they could simply state that in their positionality statement. 

We, however, recognize that authors might also be concerned about how such statements may end up disclosing axes of their identity that might negatively impact how their work is being perceived and evaluated. 
Positionality statements, however, do not necessarily need to disclose demographic or other sensitive attributes, or ``include an identity disclosure''~\citep{liang2021embracing}. 
They can instead focus on any other aspects that help the reader understand where the authors are coming from by providing clarity about the lenses they are using when conducting the work~\citep{liang2021embracing}.
As a starting point, we recommend checking the researcher guide by \cite{holmes2020researcher} and the thoughtful suggestions and examples provided by \cite{liang2021reflexivity}. 


\paragraph{2) RAI papers should include {\em ethical considerations statements}.} 
By its very nature, RAI work centers humans. 
It is thus critical that ethical considerations remain top of mind for researchers and practitioners, who should carefully consider how the individual autonomy, agency, and well-being---of e.g., those producing or represented in datasets, of those involved in (or excluded from) any other part of the research and development processes (e.g., study participants, researchers, engineers, content moderators, red team-ers), or of those expected to benefit or engage with the research outcomes---are impacted by the use of data or by how the research was conducted.  
Ethical considerations statements should especially cover ethical concerns the authors had and mitigated while conducting the work. 
These statements could include whether the authors obtained an IRB's approval for any human subject research and the concerns covered by the IRB. 
However, while IRBs to some extent set common standards and provide researchers with a framework to reflect critically about risks and benefits, and whether these risks and benefits are justly apportioned~\citep{olteanu2019social}, the ethical considerations statements should not necessarily be limited to them.

\paragraph{3) RAI papers should include {\em discussions of limitations}.} 
Reflecting on and making any data and methodological limitations explicit can further help illustrate the issues these limitations (and the resulting work) might lead to. 
Such limitations can include aspects related to research design choices such as problem framing or data and methodological choices, or aspects related to constraints that researchers need to navigate, such as access to participants, computing, or other resources. 
The discussion about limitations could, for instance, include reflections on the assumptions that a given problem framing or methodological approach makes and when those assumptions might not hold, or on the way that data biases or lack of data coverage limits the insights that can be drawn from it. 
It could also include considerations related to internal, external, or construct validity~\citep{olteanu2019social,jacobs2021measurement,blodgett2021stereotyping}. 
If the authors believe their work has no limitations, they could note this. 
While discussions of methodological limitations are more commonly included in research papers across disciplines, we believed it is worth foregrounding them here as well, even only to clarify how they are different from ethical concerns and adverse impacts.
The work by \cite{smith2022real} might provide a useful starting point for thinking about limitations. 

\paragraph{4) RAI papers should include {\em adverse impact statements}.} 
While impact statements about possible adverse impacts can be informed by the researcher positionality, ethical considerations, and discussions of methodological limitations, they are not the same. 
For example, positionality statements are important when thinking about impacts as they help contextualize how authors prioritize problems, and thus help understand possible blind-spots they might have.
In a good impact statement, authors critically reflect on not only the impact of how the work was done (which might be covered by ethical concerns), but also on the impact the work will have once it is put out into the world and used by others---e.g., work using crowd judges to label harmful content does not only raise concerns when the research is conducted (which could go under ethical considerations), but also due to possibly recommending others to do the same.  
Adverse impact statements could also include reflections of how unintended consequences could possibly be handled, including recourse mechanisms and possible checks and balances that might help identify such consequences early on. 

Decoupling the anticipation of adverse impacts (e.g., ideating about what harms our work can give rise to) from their mitigation (e.g., how can we mitigate these possible harms), might also help authors avoid conflating the two and avoid hyper-focusing only on issues they might know how to mitigate~\citep{buccinca2023aha}.
For those unfamiliar with this practice, the guide for writing the NeurIPS impact statements~\citep{ashurst2020guide} might provide a helpful starting point, along with papers that have examined such practices at NeurIPS and ACL \citep[e.g.,][]{boyarskaya2020overcoming,nanayakkara2021unpacking,liu2022examining,benotti2022ethics}. 

\smallskip
{\qquad \qquad \qquad \qquad \qquad \qquad \qquad \qquad \qquad \quad ***}
\smallskip

We recommend all four reflections and discussions to center marginalized and vulnerable communities, particularly those at the intersection of race, ethnicity, class, gender, nationality, and other characteristics that historically and at present have led to marginalization. 
For instance, a domain that historically has motivated the development of a large portion of algorithmic fairness research are technologies and algorithms used by police, prisons, and/or judicial authorities \citep[e.g.,][]{angwin2016machine}. 
Adverse impacts, ethical considerations, limitations, and researcher positionality statements are particularly critical and urgent for research motivated by, conducted in, or impacting situations in which the capacity to exercise one's rights might be diminished \citep[``low-rights situations'' described by ][]{eubanks2018automating}, especially that of marginalized and vulnerable populations such as prison inmates, heavily policed communities, migrants, and asylum seekers.
Authors should also remember that those conducting the RAI research, those developing RAI policies and practices, or those being expected to enforce RAI policies and practices (e.g., practitioners who volunteer to do RAI, red teamers, content moderators) can and often do come from more marginalized backgrounds \citep[e.g.,][]{ali2023walking}.

\section*{Concluding Reflections}

We echo the growing body of work and the calls for embracing critical, dissenting voices~\citep{matthews2022embracing,young2022confronting} and self-reflection in our own community~\citep{barocas2020not,boyarskaya2020overcoming}. 
We believe our community should do more to critically reflect on and mitigate the possible risks and harms that RAI research and work might also give rise to.  
We hope this perspective provides a starting point and some guidance on how authors of RAI research could more meaningfully engage with possible adverse impacts of their own work.

\paragraph{Adverse impact statement. }
Articulating, writing, and sharing this viewpoint is not without risks either. 
Anticipating harms or unintended consequences is hard even when guidance is provided, and researchers and practitioners often lack training and are time and resource-constrained. 
Thus, while we believe there are benefits from requiring reflections 1)~on how our backgrounds shape our work, 2)~on what ethical issues we identified while conducting the work and how we engaged with them, 3)~on the limitations of our work, and 4)~on the types of unintended consequences that the resulting work can have, we also recognize that these are {\em not a panacea} \citep[e.g.,][]{stahl2023systematic} as many factors affect by whom, whether and when adverse impacts are foreseeable~\citep{boyarskaya2020overcoming}. 
These practices might, in fact, end up just reflecting and promoting the same perspective and values of the status quo regarding what should be prioritized as those shaping the work our community does. 

There is also a risk of overwhelming researchers and practitioners with too many requirements, and thus disincentivizing them from meaningfully engaging with the task of ideating about adverse impacts, or the task of reporting on ethical concerns and limitations, or even from conducting RAI research. 
As other communities are already doing, it might be worth for our community to explore various formats for how authors could report limitations, ethical concerns, and possible adverse impacts. 

We also want to once more acknowledge that there are concerns surrounding how positionality statements could inadvertently affect marginalized researchers and practitioners. 
Authors might be affected not only by being perceived as belonging to a group but also by being assumed as not being part of a group, as ``careless requests for such statements or using them in absolutist ways that control who can and cannot do the work can cause some of the very same harms that those who request them are hoping to mitigate''~\citep{liang2021embracing}. 
There is also a risk of misguided deference where the authors' identity is used to misconstrue their position as representative of an entire marginalized group \citep[see][]{taiwo2020being}.
Positionality statements might also accidentally de-anonymize authors during peer-review if authors disclose attributes that are shared by only a few, and venues should explore how these might interact with specific anonymization requirements. 


Finally, our perspective might be failing to foresee situations where it might not be appropriate to ask authors to include some or any of the statements we highlighted earlier. RAI researchers and practitioners might also face more opposition and friction while conducting their work, and these requirements might unwittingly and unnecessarily further strain already overwhelmed researchers and practitioners.

\paragraph{Positionality statement.} 
The research, disciplinary background, and personal views of the lead author, AO, have significantly influenced this perspective, as her own work has examined how our choices of what problems to prioritize and work on, of how we do our work, and of how we interpret research results are often shaped by unstated or implicit values, norms, goals, practices, and assumptions, as well as by our own ``failures of imagination.''  
ME similarly draws from several years of efforts to bridge between different communities, particularly RAI and the recommendation and information retrieval (IR) communities, and his use of the pedagogical idea of ``scaffolding'' to model and advocate for continuous improvement in the quality of RAI work in these communities and the attention of that work to the needs and impact on marginalized communities (including shifts in his own research methods and writing).
CC is influenced by a perspective centered on computing research, computing applications, and computing education, and by the specific concerns of the FAccT conference with which they have been involved for the past five years.
JS draws from her research at the intersection of technology and human well-being where she examines the role of technologies, design choices, and values embedded in them in shifting power dynamics and improving individual and organizational well-being. In relation to the perspective presented in this article, she draws on her research on worker well-being, especially surrounding the invisible forms of labor that underlie the creation and deployment of technologies. 

\smallskip
{\qquad \qquad \qquad \qquad \qquad \qquad \qquad \qquad \qquad \quad ***}
\smallskip

While these two statements are imperfect examples of adverse impacts and researcher positionality, we hope they illustrate how even such a viewpoint can benefit from them. 
Our critique, however, very much applies to our own work as well, and even to this perspective. 
We might have also failed to recognize and highlight possible ethical concerns and limitations of both how our perspective on how ``RAI research needs impact statements too'' came together, and of what it currently covers.

\subsubsection*{Acknowledgements}
We would like to thank Alexandra Chouldechova for early discussions about impact statements for RAI research, and Reuben Binns for insightful feedback about positionality statements.


\bibliographystyle{references-format}
\bibliography{references}


\begin{thebibliography}{45}


\ifx \showCODEN    \undefined \def \showCODEN     #1{\unskip}     \fi
\ifx \showDOI      \undefined \def \showDOI       #1{#1}\fi
\ifx \showISBNx    \undefined \def \showISBNx     #1{\unskip}     \fi
\ifx \showISBNxiii \undefined \def \showISBNxiii  #1{\unskip}     \fi
\ifx \showISSN     \undefined \def \showISSN      #1{\unskip}     \fi
\ifx \showLCCN     \undefined \def \showLCCN      #1{\unskip}     \fi
\ifx \shownote     \undefined \def \shownote      #1{#1}          \fi
\ifx \showarticletitle \undefined \def \showarticletitle #1{#1}   \fi
\ifx \showURL      \undefined \def \showURL       {\relax}        \fi
\providecommand\bibfield[2]{#2}
\providecommand\bibinfo[2]{#2}
\providecommand\natexlab[1]{#1}
\providecommand\showeprint[2][]{arXiv:#2}

\bibitem[Abebe et~al\mbox{.}(2020)]%
        {abebe2020roles}
\bibfield{author}{\bibinfo{person}{Rediet Abebe}, \bibinfo{person}{Solon
  Barocas}, \bibinfo{person}{Jon Kleinberg}, \bibinfo{person}{Karen Levy},
  \bibinfo{person}{Manish Raghavan}, {and} \bibinfo{person}{David~G Robinson}.}
  \bibinfo{year}{2020}\natexlab{}.
\newblock \showarticletitle{Roles for computing in social change}. In
  \bibinfo{booktitle}{\emph{Proceedings of the 2020 conference on fairness,
  accountability, and transparency}}. \bibinfo{pages}{252--260}.
\newblock


\bibitem[Ali et~al\mbox{.}(2023)]%
        {ali2023walking}
\bibfield{author}{\bibinfo{person}{Sanna~J Ali}, \bibinfo{person}{Ang{\`e}le
  Christin}, \bibinfo{person}{Andrew Smart}, {and} \bibinfo{person}{Riitta
  Katila}.} \bibinfo{year}{2023}\natexlab{}.
\newblock \showarticletitle{Walking the Walk of AI Ethics: Organizational
  Challenges and the Individualization of Risk among Ethics Entrepreneurs}. In
  \bibinfo{booktitle}{\emph{Proceedings of the 2023 ACM Conference on Fairness,
  Accountability, and Transparency}}. \bibinfo{pages}{217--226}.
\newblock


\bibitem[Angwin et~al\mbox{.}(2016)]%
        {angwin2016machine}
\bibfield{author}{\bibinfo{person}{Julia Angwin}, \bibinfo{person}{Jeff
  Larson}, \bibinfo{person}{Surya Mattu}, {and} \bibinfo{person}{Lauren
  Kirchner}.} \bibinfo{year}{2016}\natexlab{}.
\newblock \showarticletitle{Machine bias}.
\newblock \bibinfo{journal}{\emph{ProPublica}} (\bibinfo{year}{2016}).
\newblock


\bibitem[Ashurst et~al\mbox{.}(2020)]%
        {ashurst2020guide}
\bibfield{author}{\bibinfo{person}{Carolyn Ashurst}, \bibinfo{person}{Markus
  Anderljung}, \bibinfo{person}{Carina Prunkl}, \bibinfo{person}{Jan Leike},
  \bibinfo{person}{Yarin Gal}, \bibinfo{person}{Toby Shevlane}, {and}
  \bibinfo{person}{Allan Dafoe}.} \bibinfo{year}{2020}\natexlab{}.
\newblock \showarticletitle{A guide to writing the NeurIPS impact statement}.
\newblock \bibinfo{journal}{\emph{Centre for the Governance of AI. URL:
  https://perma. cc/B5R8-2B9V}} (\bibinfo{year}{2020}).
\newblock


\bibitem[Ashurst et~al\mbox{.}(2022)]%
        {ashurst2022ai}
\bibfield{author}{\bibinfo{person}{Carolyn Ashurst}, \bibinfo{person}{Emmie
  Hine}, \bibinfo{person}{Paul Sedille}, {and} \bibinfo{person}{Alexis
  Carlier}.} \bibinfo{year}{2022}\natexlab{}.
\newblock \showarticletitle{Ai ethics statements: analysis and lessons learnt
  from neurips broader impact statements}. In
  \bibinfo{booktitle}{\emph{Proceedings of the 2022 ACM Conference on Fairness,
  Accountability, and Transparency}}. \bibinfo{pages}{2047--2056}.
\newblock


\bibitem[Balayn et~al\mbox{.}(2023)]%
        {balayn2023fairness}
\bibfield{author}{\bibinfo{person}{Agathe Balayn}, \bibinfo{person}{Mireia
  Yurrita}, \bibinfo{person}{Jie Yang}, {and} \bibinfo{person}{Ujwal
  Gadiraju}.} \bibinfo{year}{2023}\natexlab{}.
\newblock \showarticletitle{``Fairness Toolkits, A Checkbox Culture?'' On the
  Factors that Fragment Developer Practices in Handling Algorithmic Harms}. In
  \bibinfo{booktitle}{\emph{Proceedings of the 2023 AAAI/ACM Conference on AI,
  Ethics, and Society}}. \bibinfo{pages}{482--495}.
\newblock


\bibitem[Barocas et~al\mbox{.}(2020)]%
        {barocas2020not}
\bibfield{author}{\bibinfo{person}{Solon Barocas}, \bibinfo{person}{Asia~J
  Biega}, \bibinfo{person}{Benjamin Fish}, \bibinfo{person}{J{\k{e}}drzej
  Niklas}, {and} \bibinfo{person}{Luke Stark}.}
  \bibinfo{year}{2020}\natexlab{}.
\newblock \showarticletitle{When not to design, build, or deploy}. In
  \bibinfo{booktitle}{\emph{Proceedings of the 2020 Conference on Fairness,
  Accountability, and Transparency}}. \bibinfo{pages}{695--695}.
\newblock


\bibitem[Bennett and Keyes(2020)]%
        {bennett2020point}
\bibfield{author}{\bibinfo{person}{Cynthia~L Bennett} {and} \bibinfo{person}{Os
  Keyes}.} \bibinfo{year}{2020}\natexlab{}.
\newblock \showarticletitle{What is the point of fairness? Disability, AI and
  the complexity of justice}.
\newblock \bibinfo{journal}{\emph{ACM SIGACCESS Accessibility and Computing}}
  \bibinfo{number}{125} (\bibinfo{year}{2020}), \bibinfo{pages}{1--1}.
\newblock


\bibitem[Benotti and Blackburn(2022)]%
        {benotti2022ethics}
\bibfield{author}{\bibinfo{person}{Luciana Benotti} {and}
  \bibinfo{person}{Patrick Blackburn}.} \bibinfo{year}{2022}\natexlab{}.
\newblock \showarticletitle{Ethics consideration sections in natural language
  processing papers}. In \bibinfo{booktitle}{\emph{Proceedings of the 2022
  Conference on Empirical Methods in Natural Language Processing}}.
  \bibinfo{pages}{4509--4516}.
\newblock


\bibitem[Birhane et~al\mbox{.}(2022)]%
        {birhane2022forgotten}
\bibfield{author}{\bibinfo{person}{Abeba Birhane}, \bibinfo{person}{Elayne
  Ruane}, \bibinfo{person}{Thomas Laurent}, \bibinfo{person}{Matthew S.~Brown},
  \bibinfo{person}{Johnathan Flowers}, \bibinfo{person}{Anthony Ventresque},
  {and} \bibinfo{person}{Christopher L.~Dancy}.}
  \bibinfo{year}{2022}\natexlab{}.
\newblock \showarticletitle{The forgotten margins of AI ethics}. In
  \bibinfo{booktitle}{\emph{Proceedings of the 2022 ACM Conference on Fairness,
  Accountability, and Transparency}}. \bibinfo{pages}{948--958}.
\newblock


\bibitem[Blodgett et~al\mbox{.}(2021)]%
        {blodgett2021stereotyping}
\bibfield{author}{\bibinfo{person}{Su~Lin Blodgett}, \bibinfo{person}{Gilsinia
  Lopez}, \bibinfo{person}{Alexandra Olteanu}, \bibinfo{person}{Robert Sim},
  {and} \bibinfo{person}{Hanna Wallach}.} \bibinfo{year}{2021}\natexlab{}.
\newblock \showarticletitle{Stereotyping Norwegian salmon: An inventory of
  pitfalls in fairness benchmark datasets}. In
  \bibinfo{booktitle}{\emph{Proceedings of the 59th Annual Meeting of the
  Association for Computational Linguistics and the 11th International Joint
  Conference on Natural Language Processing (Volume 1: Long Papers)}}.
  \bibinfo{pages}{1004--1015}.
\newblock


\bibitem[Boyarskaya et~al\mbox{.}(2020)]%
        {boyarskaya2020overcoming}
\bibfield{author}{\bibinfo{person}{Margarita Boyarskaya},
  \bibinfo{person}{Alexandra Olteanu}, {and} \bibinfo{person}{Kate Crawford}.}
  \bibinfo{year}{2020}\natexlab{}.
\newblock \showarticletitle{Overcoming failures of imagination in AI infused
  system development and deployment}.
\newblock \bibinfo{journal}{\emph{arXiv preprint arXiv:2011.13416}}
  (\bibinfo{year}{2020}).
\newblock


\bibitem[Bu{\c{c}}inca et~al\mbox{.}(2023)]%
        {buccinca2023aha}
\bibfield{author}{\bibinfo{person}{Zana Bu{\c{c}}inca},
  \bibinfo{person}{Chau~Minh Pham}, \bibinfo{person}{Maurice Jakesch},
  \bibinfo{person}{Marco~Tulio Ribeiro}, \bibinfo{person}{Alexandra Olteanu},
  {and} \bibinfo{person}{Saleema Amershi}.} \bibinfo{year}{2023}\natexlab{}.
\newblock \showarticletitle{AHA!: Facilitating AI Impact Assessment by
  Generating Examples of Harms}.
\newblock \bibinfo{journal}{\emph{arXiv preprint arXiv:2306.03280}}
  (\bibinfo{year}{2023}).
\newblock


\bibitem[Eubanks(2018)]%
        {eubanks2018automating}
\bibfield{author}{\bibinfo{person}{Virginia Eubanks}.}
  \bibinfo{year}{2018}\natexlab{}.
\newblock \bibinfo{booktitle}{\emph{Automating inequality: How high-tech tools
  profile, police, and punish the poor}}.
\newblock \bibinfo{publisher}{St. Martin's Press}.
\newblock


\bibitem[Gansky and McDonald(2022)]%
        {gansky2022counterfacctual}
\bibfield{author}{\bibinfo{person}{Ben Gansky} {and} \bibinfo{person}{Sean
  McDonald}.} \bibinfo{year}{2022}\natexlab{}.
\newblock \showarticletitle{CounterFAccTual: How FAccT undermines its
  organizing principles}. In \bibinfo{booktitle}{\emph{Proceedings of the 2022
  ACM Conference on Fairness, Accountability, and Transparency}}.
  \bibinfo{pages}{1982--1992}.
\newblock


\bibitem[Green and Viljoen(2020)]%
        {green2020algorithmic}
\bibfield{author}{\bibinfo{person}{Ben Green} {and} \bibinfo{person}{Salom{\'e}
  Viljoen}.} \bibinfo{year}{2020}\natexlab{}.
\newblock \showarticletitle{Algorithmic realism: expanding the boundaries of
  algorithmic thought}. In \bibinfo{booktitle}{\emph{Proceedings of the 2020
  conference on fairness, accountability, and transparency}}.
  \bibinfo{pages}{19--31}.
\newblock


\bibitem[Hecht(2020)]%
        {hecht2020suggestions}
\bibfield{author}{\bibinfo{person}{Brent Hecht}.}
  \bibinfo{year}{2020}\natexlab{}.
\newblock \showarticletitle{Suggestions for Writing NeurIPS 2020 Broader
  Impacts Statements}.
\newblock \bibinfo{journal}{\emph{Medium. https://medium. com/@
  BrentH/suggestions-for-writing-neurips-2020-broader-impacts-statements-121da1b765bf}}
  (\bibinfo{year}{2020}).
\newblock


\bibitem[Hecht et~al\mbox{.}(2018)]%
        {hecht2021s}
\bibfield{author}{\bibinfo{person}{Brent Hecht}, \bibinfo{person}{Lauren
  Wilcox}, \bibinfo{person}{Jeffrey~P Bigham}, \bibinfo{person}{Johannes
  Sch{\"o}ning}, \bibinfo{person}{Ehsan Hoque}, \bibinfo{person}{Jason Ernst},
  \bibinfo{person}{Yonatan Bisk}, \bibinfo{person}{Luigi De~Russis},
  \bibinfo{person}{Lana Yarosh}, \bibinfo{person}{Bushra Anjum},
  {et~al\mbox{.}}} \bibinfo{year}{2018}\natexlab{}.
\newblock \showarticletitle{It's time to do something: Mitigating the negative
  impacts of computing through a change to the peer review process}.
\newblock \bibinfo{journal}{\emph{ACM Future of Computing Blog}}
  (\bibinfo{year}{2018}).
\newblock


\bibitem[Heikkil{\"a}(2022)]%
        {heikkila2022responsible}
\bibfield{author}{\bibinfo{person}{Melissa Heikkil{\"a}}.}
  \bibinfo{year}{2022}\natexlab{}.
\newblock \showarticletitle{Responsible AI has a burnout problem}.
\newblock \bibinfo{journal}{\emph{MIT Technology Review. October}}
  \bibinfo{volume}{28} (\bibinfo{year}{2022}), \bibinfo{pages}{2022}.
\newblock


\bibitem[Holmes(2020)]%
        {holmes2020researcher}
\bibfield{author}{\bibinfo{person}{Andrew Gary~Darwin Holmes}.}
  \bibinfo{year}{2020}\natexlab{}.
\newblock \showarticletitle{Researcher Positionality--A Consideration of Its
  Influence and Place in Qualitative Research--A New Researcher Guide.}
\newblock \bibinfo{journal}{\emph{Shanlax International Journal of Education}}
  \bibinfo{volume}{8}, \bibinfo{number}{4} (\bibinfo{year}{2020}),
  \bibinfo{pages}{1--10}.
\newblock


\bibitem[Jacobs and Wallach(2021)]%
        {jacobs2021measurement}
\bibfield{author}{\bibinfo{person}{Abigail~Z Jacobs} {and}
  \bibinfo{person}{Hanna Wallach}.} \bibinfo{year}{2021}\natexlab{}.
\newblock \showarticletitle{Measurement and fairness}. In
  \bibinfo{booktitle}{\emph{Proceedings of the 2021 ACM conference on fairness,
  accountability, and transparency}}. \bibinfo{pages}{375--385}.
\newblock


\bibitem[Jakesch et~al\mbox{.}(2022)]%
        {jakesch2022different}
\bibfield{author}{\bibinfo{person}{Maurice Jakesch}, \bibinfo{person}{Zana
  Bu{\c{c}}inca}, \bibinfo{person}{Saleema Amershi}, {and}
  \bibinfo{person}{Alexandra Olteanu}.} \bibinfo{year}{2022}\natexlab{}.
\newblock \showarticletitle{How different groups prioritize ethical values for
  responsible AI}. In \bibinfo{booktitle}{\emph{Proceedings of the 2022 ACM
  Conference on Fairness, Accountability, and Transparency}}.
  \bibinfo{pages}{310--323}.
\newblock


\bibitem[Keyes et~al\mbox{.}(2019)]%
        {keyes2019mulching}
\bibfield{author}{\bibinfo{person}{Os Keyes}, \bibinfo{person}{Jevan Hutson},
  {and} \bibinfo{person}{Meredith Durbin}.} \bibinfo{year}{2019}\natexlab{}.
\newblock \showarticletitle{A mulching proposal: Analysing and improving an
  algorithmic system for turning the elderly into high-nutrient slurry}. In
  \bibinfo{booktitle}{\emph{Extended abstracts of the 2019 CHI conference on
  human factors in computing systems}}. \bibinfo{pages}{1--11}.
\newblock


\bibitem[Laufer et~al\mbox{.}(2022)]%
        {laufer2022four}
\bibfield{author}{\bibinfo{person}{Benjamin Laufer}, \bibinfo{person}{Sameer
  Jain}, \bibinfo{person}{A~Feder Cooper}, \bibinfo{person}{Jon Kleinberg},
  {and} \bibinfo{person}{Hoda Heidari}.} \bibinfo{year}{2022}\natexlab{}.
\newblock \showarticletitle{Four years of FAccT: A reflexive, mixed-methods
  analysis of research contributions, shortcomings, and future prospects}. In
  \bibinfo{booktitle}{\emph{2022 ACM Conference on Fairness, Accountability,
  and Transparency}}. \bibinfo{pages}{401--426}.
\newblock


\bibitem[Liang(2021)]%
        {liang2021reflexivity}
\bibfield{author}{\bibinfo{person}{Calvin Liang}.}
  \bibinfo{year}{2021}\natexlab{}.
\newblock \bibinfo{title}{Reflexivity, positionality, and disclosure in HCI}.
\newblock
\newblock


\bibitem[Liang et~al\mbox{.}(2021)]%
        {liang2021embracing}
\bibfield{author}{\bibinfo{person}{Calvin~A Liang}, \bibinfo{person}{Sean~A
  Munson}, {and} \bibinfo{person}{Julie~A Kientz}.}
  \bibinfo{year}{2021}\natexlab{}.
\newblock \showarticletitle{Embracing four tensions in human-computer
  interaction research with marginalized people}.
\newblock \bibinfo{journal}{\emph{ACM Transactions on Computer-Human
  Interaction (TOCHI)}} \bibinfo{volume}{28}, \bibinfo{number}{2}
  (\bibinfo{year}{2021}), \bibinfo{pages}{1--47}.
\newblock


\bibitem[Liu et~al\mbox{.}(2022)]%
        {liu2022examining}
\bibfield{author}{\bibinfo{person}{David Liu}, \bibinfo{person}{Priyanka
  Nanayakkara}, \bibinfo{person}{Sarah~Ariyan Sakha}, \bibinfo{person}{Grace
  Abuhamad}, \bibinfo{person}{Su~Lin Blodgett}, \bibinfo{person}{Nicholas
  Diakopoulos}, \bibinfo{person}{Jessica~R Hullman}, {and}
  \bibinfo{person}{Tina Eliassi-Rad}.} \bibinfo{year}{2022}\natexlab{}.
\newblock \showarticletitle{Examining Responsibility and Deliberation in AI
  Impact Statements and Ethics Reviews}. In
  \bibinfo{booktitle}{\emph{Proceedings of the 2022 AAAI/ACM Conference on AI,
  Ethics, and Society}}. \bibinfo{pages}{424--435}.
\newblock


\bibitem[Matthews(2022)]%
        {matthews2022embracing}
\bibfield{author}{\bibinfo{person}{Jeanna Matthews}.}
  \bibinfo{year}{2022}\natexlab{}.
\newblock \bibinfo{title}{Embracing critical voices}.
\newblock , \bibinfo{numpages}{7--7}~pages.
\newblock


\bibitem[Nanayakkara et~al\mbox{.}(2021)]%
        {nanayakkara2021unpacking}
\bibfield{author}{\bibinfo{person}{Priyanka Nanayakkara},
  \bibinfo{person}{Jessica Hullman}, {and} \bibinfo{person}{Nicholas
  Diakopoulos}.} \bibinfo{year}{2021}\natexlab{}.
\newblock \showarticletitle{Unpacking the expressed consequences of AI research
  in broader impact statements}. In \bibinfo{booktitle}{\emph{Proceedings of
  the 2021 AAAI/ACM Conference on AI, Ethics, and Society}}.
  \bibinfo{pages}{795--806}.
\newblock


\bibitem[Olteanu et~al\mbox{.}(2019)]%
        {olteanu2019social}
\bibfield{author}{\bibinfo{person}{Alexandra Olteanu}, \bibinfo{person}{Carlos
  Castillo}, \bibinfo{person}{Fernando Diaz}, {and} \bibinfo{person}{Emre
  K{\i}c{\i}man}.} \bibinfo{year}{2019}\natexlab{}.
\newblock \showarticletitle{Social data: Biases, methodological pitfalls, and
  ethical boundaries}.
\newblock \bibinfo{journal}{\emph{Frontiers in big data}}  \bibinfo{volume}{2}
  (\bibinfo{year}{2019}), \bibinfo{pages}{13}.
\newblock


\bibitem[Pinney et~al\mbox{.}(2023)]%
        {pinneyMuchAdoGender2023}
\bibfield{author}{\bibinfo{person}{Christine Pinney}, \bibinfo{person}{Amifa
  Raj}, \bibinfo{person}{Alex Hanna}, {and} \bibinfo{person}{Michael~D
  Ekstrand}.} \bibinfo{year}{2023}\natexlab{}.
\newblock \showarticletitle{Much {{Ado About Gender}}: {{Current Practices}}
  and {{Future Recommendations}} for {{Appropriate Gender-Aware Information
  Access}}}. In \bibinfo{booktitle}{\emph{{{CHIIR}} '23}}.
  \bibinfo{publisher}{{Association for Computing Machinery}},
  \bibinfo{address}{{New York, NY, USA}}, \bibinfo{pages}{269--279}.
\newblock


\bibitem[Prunkl et~al\mbox{.}(2021)]%
        {prunkl2021institutionalizing}
\bibfield{author}{\bibinfo{person}{Carina~EA Prunkl}, \bibinfo{person}{Carolyn
  Ashurst}, \bibinfo{person}{Markus Anderljung}, \bibinfo{person}{Helena Webb},
  \bibinfo{person}{Jan Leike}, {and} \bibinfo{person}{Allan Dafoe}.}
  \bibinfo{year}{2021}\natexlab{}.
\newblock \showarticletitle{Institutionalizing ethics in AI through broader
  impact requirements}.
\newblock \bibinfo{journal}{\emph{Nature Machine Intelligence}}
  \bibinfo{volume}{3}, \bibinfo{number}{2} (\bibinfo{year}{2021}),
  \bibinfo{pages}{104--110}.
\newblock


\bibitem[Raji et~al\mbox{.}(2022)]%
        {raji2022fallacy}
\bibfield{author}{\bibinfo{person}{Inioluwa~Deborah Raji},
  \bibinfo{person}{I~Elizabeth Kumar}, \bibinfo{person}{Aaron Horowitz}, {and}
  \bibinfo{person}{Andrew Selbst}.} \bibinfo{year}{2022}\natexlab{}.
\newblock \showarticletitle{The fallacy of AI functionality}. In
  \bibinfo{booktitle}{\emph{2022 ACM Conference on Fairness, Accountability,
  and Transparency}}. \bibinfo{pages}{959--972}.
\newblock


\bibitem[Rakova et~al\mbox{.}(2021)]%
        {rakova2021responsible}
\bibfield{author}{\bibinfo{person}{Bogdana Rakova}, \bibinfo{person}{Jingying
  Yang}, \bibinfo{person}{Henriette Cramer}, {and} \bibinfo{person}{Rumman
  Chowdhury}.} \bibinfo{year}{2021}\natexlab{}.
\newblock \showarticletitle{Where responsible AI meets reality: Practitioner
  perspectives on enablers for shifting organizational practices}.
\newblock \bibinfo{journal}{\emph{Proceedings of the ACM on Human-Computer
  Interaction}} \bibinfo{volume}{5}, \bibinfo{number}{CSCW1}
  (\bibinfo{year}{2021}), \bibinfo{pages}{1--23}.
\newblock


\bibitem[Robertson et~al\mbox{.}(2021)]%
        {robertson2021can}
\bibfield{author}{\bibinfo{person}{Ronald~E Robertson},
  \bibinfo{person}{Alexandra Olteanu}, \bibinfo{person}{Fernando Diaz},
  \bibinfo{person}{Milad Shokouhi}, {and} \bibinfo{person}{Peter Bailey}.}
  \bibinfo{year}{2021}\natexlab{}.
\newblock \showarticletitle{“I can’t reply with that”: Characterizing
  problematic email reply suggestions}. In
  \bibinfo{booktitle}{\emph{Proceedings of the 2021 CHI Conference on Human
  Factors in Computing Systems}}. \bibinfo{pages}{1--18}.
\newblock


\bibitem[Sandvig et~al\mbox{.}(2014)]%
        {sandvig2014auditing}
\bibfield{author}{\bibinfo{person}{Christian Sandvig}, \bibinfo{person}{Kevin
  Hamilton}, \bibinfo{person}{Karrie Karahalios}, {and} \bibinfo{person}{Cedric
  Langbort}.} \bibinfo{year}{2014}\natexlab{}.
\newblock \showarticletitle{Auditing algorithms: Research methods for detecting
  discrimination on internet platforms}.
\newblock \bibinfo{journal}{\emph{Data and discrimination: converting critical
  concerns into productive inquiry}} \bibinfo{volume}{22},
  \bibinfo{number}{2014} (\bibinfo{year}{2014}), \bibinfo{pages}{4349--4357}.
\newblock


\bibitem[Septiandri et~al\mbox{.}(2023)]%
        {septiandri2023weird}
\bibfield{author}{\bibinfo{person}{Ali~Akbar Septiandri},
  \bibinfo{person}{Marios Constantinides}, \bibinfo{person}{Mohammad Tahaei},
  {and} \bibinfo{person}{Daniele Quercia}.} \bibinfo{year}{2023}\natexlab{}.
\newblock \showarticletitle{WEIRD FAccTs: How Western, Educated,
  Industrialized, Rich, and Democratic is FAccT?}. In
  \bibinfo{booktitle}{\emph{Proceedings of the 2023 ACM Conference on Fairness,
  Accountability, and Transparency}}. \bibinfo{pages}{160--171}.
\newblock


\bibitem[Smith et~al\mbox{.}(2022)]%
        {smith2022real}
\bibfield{author}{\bibinfo{person}{Jessie~J Smith}, \bibinfo{person}{Saleema
  Amershi}, \bibinfo{person}{Solon Barocas}, \bibinfo{person}{Hanna Wallach},
  {and} \bibinfo{person}{Jennifer Wortman~Vaughan}.}
  \bibinfo{year}{2022}\natexlab{}.
\newblock \showarticletitle{Real ml: Recognizing, exploring, and articulating
  limitations of machine learning research}. In
  \bibinfo{booktitle}{\emph{Proceedings of the 2022 ACM Conference on Fairness,
  Accountability, and Transparency}}. \bibinfo{pages}{587--597}.
\newblock


\bibitem[Stahl et~al\mbox{.}(2023)]%
        {stahl2023systematic}
\bibfield{author}{\bibinfo{person}{Bernd~Carsten Stahl},
  \bibinfo{person}{Josephina Antoniou}, \bibinfo{person}{Nitika Bhalla},
  \bibinfo{person}{Laurence Brooks}, \bibinfo{person}{Philip Jansen},
  \bibinfo{person}{Blerta Lindqvist}, \bibinfo{person}{Alexey Kirichenko},
  \bibinfo{person}{Samuel Marchal}, \bibinfo{person}{Rowena Rodrigues},
  \bibinfo{person}{Nicole Santiago}, {et~al\mbox{.}}}
  \bibinfo{year}{2023}\natexlab{}.
\newblock \showarticletitle{A systematic review of artificial intelligence
  impact assessments}.
\newblock \bibinfo{journal}{\emph{Artificial Intelligence Review}}
  (\bibinfo{year}{2023}), \bibinfo{pages}{1--33}.
\newblock


\bibitem[T\'a\'iw\`o(2020)]%
        {taiwo2020being}
\bibfield{author}{\bibinfo{person}{Ol\'uf\d{\'e}mi T\'a\'iw\`o}.}
  \bibinfo{year}{2020}\natexlab{}.
\newblock \showarticletitle{Being-in-the-room privilege: Elite capture and
  epistemic deference}.
\newblock \bibinfo{journal}{\emph{The Philosopher}} \bibinfo{volume}{108},
  \bibinfo{number}{4} (\bibinfo{year}{2020}), \bibinfo{pages}{61--70}.
\newblock


\bibitem[Widder et~al\mbox{.}(2023)]%
        {widder2023s}
\bibfield{author}{\bibinfo{person}{David~Gray Widder}, \bibinfo{person}{Derrick
  Zhen}, \bibinfo{person}{Laura Dabbish}, {and} \bibinfo{person}{James
  Herbsleb}.} \bibinfo{year}{2023}\natexlab{}.
\newblock \showarticletitle{It’s about power: What ethical concerns do
  software engineers have, and what do they (feel they can) do about them?}. In
  \bibinfo{booktitle}{\emph{Proceedings of the 2023 ACM Conference on Fairness,
  Accountability, and Transparency}}. \bibinfo{pages}{467--479}.
\newblock


\bibitem[Wilkinson et~al\mbox{.}(2023)]%
        {craftTOC2023}
\bibfield{author}{\bibinfo{person}{Daricia Wilkinson},
  \bibinfo{person}{{Michael Ekstrand}}, \bibinfo{person}{{Janet A. Vertesi}},
  {and} \bibinfo{person}{{Alexandra Olteanu}}.}
  \bibinfo{year}{2023}\natexlab{}.
\newblock \bibinfo{title}{Theories of Change in Responsible AI. {\em CRAFT
  Session at the 2023 Conference on Fairness, Accountability, and
  Transparency.}}
\newblock
\newblock


\bibitem[Wong et~al\mbox{.}(2023)]%
        {wong2023seeing}
\bibfield{author}{\bibinfo{person}{Richmond~Y Wong}, \bibinfo{person}{Michael~A
  Madaio}, {and} \bibinfo{person}{Nick Merrill}.}
  \bibinfo{year}{2023}\natexlab{}.
\newblock \showarticletitle{Seeing like a toolkit: How toolkits envision the
  work of AI ethics}.
\newblock \bibinfo{journal}{\emph{Proceedings of the ACM on Human-Computer
  Interaction}} \bibinfo{volume}{7}, \bibinfo{number}{CSCW1}
  (\bibinfo{year}{2023}), \bibinfo{pages}{1--27}.
\newblock


\bibitem[Young et~al\mbox{.}(2022)]%
        {young2022confronting}
\bibfield{author}{\bibinfo{person}{Meg Young}, \bibinfo{person}{Michael
  Katell}, {and} \bibinfo{person}{PM Krafft}.} \bibinfo{year}{2022}\natexlab{}.
\newblock \showarticletitle{Confronting power and corporate capture at the
  FAccT Conference}. In \bibinfo{booktitle}{\emph{Proceedings of the 2022 ACM
  Conference on Fairness, Accountability, and Transparency}}.
  \bibinfo{pages}{1375--1386}.
\newblock


\bibitem[Zhou et~al\mbox{.}(2022)]%
        {zhou2022deconstructing}
\bibfield{author}{\bibinfo{person}{Kaitlyn Zhou}, \bibinfo{person}{Su~Lin
  Blodgett}, \bibinfo{person}{Adam Trischler}, \bibinfo{person}{Hal
  Daum{\'e}~III}, \bibinfo{person}{Kaheer Suleman}, {and}
  \bibinfo{person}{Alexandra Olteanu}.} \bibinfo{year}{2022}\natexlab{}.
\newblock \showarticletitle{Deconstructing NLG Evaluation: Evaluation
  Practices, Assumptions, and Their Implications}. In
  \bibinfo{booktitle}{\emph{Proceedings of the 2022 Conference of the North
  American Chapter of the Association for Computational Linguistics: Human
  Language Technologies}}. \bibinfo{pages}{314--324}.
\newblock


\end{thebibliography}

\end{document}